\documentclass[conference]{IEEEtran} 
\IEEEoverridecommandlockouts

\newcommand{\figureref}[1]{Fig.~\ref{#1}}
\newcommand{\tableref}[1]{Tab.~\ref{#1}}

\usepackage{subfigure}

\usepackage{cite}
\usepackage{amsmath,amssymb,amsfonts}
\usepackage{algorithmic}
\usepackage{graphicx}
\usepackage{textcomp}
\usepackage{xcolor}
\def\BibTeX{{\rm B\kern-.05em{\sc i\kern-.025em b}\kern-.08em
    T\kern-.1667em\lower.7ex\hbox{E}\kern-.125emX}}
    
\usepackage{booktabs}
\usepackage{standalone}
\usepackage{subfigure}
\usepackage{bbm} % for \mathbbm{1}_{condition}
\usepackage{tikz} % for tikz folder imports
\usepackage{pgfplots}
\pgfplotsset{compat=newest}
\usetikzlibrary{matrix,calc,fit,positioning,arrows, arrows.meta}

\usepackage{siunitx}

\usepackage{multirow}

\usepackage{etoolbox}
\makeatletter
\patchcmd{\@makecaption}
  {\scshape}
  {}
  {}
  {}
  
\usepackage{stfloats}
\fnbelowfloat %  
  
\newcommand{\support}{SUPPORT}
\newcommand{\metabric}{METABRIC}
\newcommand{\flchain}{FLCHAIN}

\newcommand{\coxph}{CoxPH}
\newcommand{\deepsurv}{DeepSurv}
\newcommand{\coxtime}{CoxTime}
\newcommand{\drsa}{DRSA}
\newcommand{\kamran}{Kamran}
\newcommand{\dcs}{DCS}

\newcommand{\cindex}{Concordance Index}
\newcommand{\cindexabb}{C-index}
\newcommand{\cindextd}{Time-Dependent Concordance Index}
\newcommand{\cindextdabb}{C-index-td}

\newcommand{\cdauc}{Cumulative-Dynamic AUROC}
\newcommand{\cdaucabb}{CDAUC}
\newcommand{\ddc}{Distributional Divergence for Calibration}
\newcommand{\ddcabb}{DDC}

\newcommand{\lrps}{$\mathcal{L}_{\text{RPS}}$}
\newcommand{\lkernel}{$\mathcal{L}_{\text{kernel}}$}
\newcommand{\lkernelours}{$\mathcal{\tilde{L}}_{kernel}$}

\usepackage{xr-hyper}
\usepackage{hyperref}

\makeatletter
\newcommand*{\addFileDependency}[1]{% argument=file name and extension
  \typeout{(#1)}
  \@addtofilelist{#1}
  \IfFileExists{#1}{}{\typeout{No file #1.}}
}
\makeatother

\newcommand*{\myexternaldocument}[1]{%
    \externaldocument{#1}%
    \addFileDependency{#1.tex}%
    \addFileDependency{#1.aux}%
}

\myexternaldocument{supplemental-material}

\begin{document}
\bstctlcite{IEEEexample:BSTcontrol}
 % IEEE HEADER
 
% Header

\title{Deep Learning-Based Discrete Calibrated Survival Prediction}

\author{\IEEEauthorblockN{Patrick Fuhlert, Anne Ernst, Esther Dietrich, Fabian Westhaeusser, Karin Kloiber, Stefan Bonn}
\IEEEauthorblockA{\textit{Institute of Medical Systems Biology, Center for Biomedical AI (bAIome), Center for Molecular Neurobiology (ZMNH)}\\
\textit{University Medical Center Hamburg-Eppendorf}, Hamburg, Germany}
}

 % END IEEE HEADER

{\maketitle}

% to get page numbers:
% \thispagestyle{plain}
% \pagestyle{plain}

\begin{abstract}
Deep neural networks for survival prediction outperform classical approaches in discrimination, which is the ordering of patients according to their time-of-event. Conversely, classical approaches like the Cox Proportional Hazards model display much better calibration, the correct temporal prediction of events of the underlying distribution. Especially in the medical domain, where it is critical to predict the survival of a single patient, both discrimination and calibration are important performance metrics. Here we present \textbf{D}iscrete \textbf{C}alibrated \textbf{S}urvival (DCS), a novel deep neural network for discriminated and calibrated survival prediction that outperforms competing survival models in discrimination on three medical datasets, while achieving best calibration among all discrete time models. The enhanced performance of \dcs{} can be attributed to two novel features, the variable temporal output node spacing and the novel loss term that optimizes the use of uncensored and censored patient data. We believe that DCS is an important step towards clinical application of deep-learning-based survival prediction with state-of-the-art discrimination and good calibration.
\end{abstract}

\begin{IEEEkeywords}
deep learning, discrimination, calibration, survival analysis, patient records, personalized medicine, clinical decision support, electronic health record
\end{IEEEkeywords}

\section{Introduction}
Survival analysis or, in more general terms, time-to-event analysis, is a branch of statistics that analyses lifetimes of individuals or populations  regarding a particular event (e.g. death) and infers what determines the underlying distributions \cite{Klein2003}. In biomedical statistics, longitudinal patient data is analyzed to predict, for instance, the impact of a specific treatment on patient survival. This population-level analysis can determine which (groups of) patients benefit most from a given treatment and which prognostic features might be responsible for this. Classical statistics-based approaches for survival analysis include Kaplan-Meier (KM) survival curve analysis \cite{Kaplan1958} and the Cox Proportional Hazards model (CoxPH) \cite{Cox1972}. While KM analysis can be used to compare different patient groups characterized by one or more properties, the CoxPH model computes the relative risk of an event with respect to a reference population, on the basis of patient features or diagnostic results. Both approaches account for the characteristic right-censoring of patients, which refers to the fact that patients drop out due to unidentified reasons during the study. The use of this partial information of censored individuals distinguishes survival analysis from regression problems \cite{Meng2021}. Other forms of censoring are rarely encountered in medical patient data, wherefore we focus on right-censored data in this study. The predictive performance of survival models is commonly evaluated with the \cindex{} (\cindexabb{}) and \cdauc{} (\cdaucabb{}), which are measures of discrimination that assess the correct ranking of predicted risks. While discrimination quantifies if a model predicts the correct order of events, it does not take into account if the predicted event occurs at the correct time. Especially in a clinical setting however, the correct prediction of the event time is of high relevance for the patient and medical practitioner, as it guides their decision making. The correct temporal prediction of an event can be measured by the calibration of a model, for example the \ddc{} (\ddcabb{}) \cite{Haider2018}. Recent work concentrates on the development of deep learning-based (DL) algorithms for survival prediction, including models that produce a continuous time output like DeepSurv \cite{Katzman2018} and CoxTime \cite{Kvamme2019}. Recent DL approaches such as \drsa{} \cite{Ren2019} focus on the prediction of risk at discrete time points. While these discrete DL models reach state-of-the-art discrimination performance, they disregard model calibration, which makes them of limited value for medical survival prediction. To address this shortcoming, \kamran{} \textit{et al.} developed a discrete DL model that focuses on model calibration \cite{Kamran2021}. In this work we present \textbf{D}iscrete \textbf{C}alibrated \textbf{S}urvival (\dcs{}), a survival model that extends the work of \drsa{} and \kamran{} to reach state-of-the-art discrimination while being well calibrated. We show that \dcs{} outperforms five competing survival models, including \drsa{} and \kamran{}, in discrimination on three public medical tabular datasets, while achieving the best overall calibration for all discrete-time models. The performance gain in discrimination (\cdaucabb{}{} of $0.657$ on \support{}, $0.773$ on \metabric{}, $0.832$ on \flchain{}) of \dcs{} can be attributed to two novel ideas. First, \dcs{} features a modified \drsa{} architecture that allows for variable temporal output node spacing in time. Best results were obtained with quantile spacing that ensures a uniform distribution of censoring and real events per discrete time step during model training. Second, \dcs{} features a novel loss term that optimizes the use of uncensored and censored patient data to boost discrimination performance. This work uses the following structure. It starts with an overview of survival analysis and related work in \autoref{sec:theory}, followed by \autoref{sec:experiments} that shows our experimental setup. Afterwards, the results are presented in \autoref{sec:results} including comparisons to our baselines.
\section{Theory}
\label{sec:theory}
\subsection{Survival Analysis}
\begin{figure}[b]
\centerline{\includegraphics[width=\linewidth]{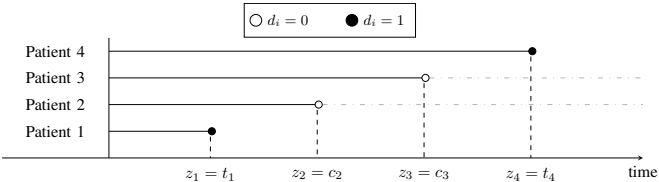}}
\caption{In time-to-event analysis, individuals either experience the event ($d_i = 1$) or are censored ($d_i = 0$) at some point in time ($z_i$).}
\label{fig:censoring}
\end{figure}

Survival analysis focuses on the correct prediction of a future event, for instance the death of a patient or the relapse of a disease. While for some patients the event of interest might be recorded, other patients might leave a given study at any point in time without actually experiencing the event, giving rise to uncensored and (right-) censored patient data, respectively. Let $z_i = \min(t_i, c_i)$ be the time where an individual is either censored ($c_i$) or experiences the event of interest ($t_i$) (\figureref{fig:censoring}). From here on, let $d_i$ denote the binary event indicator for each individual $i: d_i = \mathbbm{1}_{z_i = t_i}$.
\subsection{Models}
\label{sec:models}
This section describes current state-of-the-art survival prediction models that we compare to in this work and introduces our new DCS model. %Table \ref{table:model:comparison} gives a general overview of the analyzed models.
\subsubsection{\coxph{}}
The classical approach for survival analysis is the \coxph{} model \cite{Cox1972}. It finds the parameters that best reproduce the time-ordering of the individual events by fitting a logistic regression model that is based on the \emph{Proportional Hazards Assumption} (that can be checked using Schoenfeld Residuals \cite{park2015}) and assumes \emph{linear} and \emph{independent} covariates. The features of each individual are linearly combined $g_{\pmb{\beta}}(\mathbf{x}_{i}) = \pmb{\beta}^T \mathbf{x}_{i}$ and optimized using a partial likelihood approach. Afterwards, the regression model is joined with a baseline hazard that is equal for the whole population. By design, predicted survival curves have the same shape for all individuals (effectively discarding individual temporal information) of the population and only differ in a unique parametric scaling factor per patient that depends on the individual's input features.
\subsubsection{\deepsurv{}}
Since neural networks rather learn the underlying functional dependencies than assuming them, they permit the discovery of non-linear dependencies between the survival model's covariates \cite{Katzman2018}. \deepsurv{} retains the partial likelihood function, such that all predicted survival curves have the same shape with a different scaling factor, just as in the CoxPH model. In other words, the encoding of the predictor $g(\mathbf{x}_{i})$ is calculated by a neural network that obtains the covariate information as input.
%CoxTime
\subsubsection{\coxtime{}}
Similar to \deepsurv{}, \coxtime{} \cite{Kvamme2019} uses a discriminative partial likelihood objective function, but introduces a time-variant degree of freedom in the survival curves by feeding the desired prediction time into the network as an additional input. This relaxes the proportionality constraint from the model to allow for potentially crossing survival curves that violate the proportional hazards assumption. 
%DRSA
\subsubsection{\drsa{}}
The algorithm abandons the log-likelihood loss and instead computes the conditional probabilities of the event over time \cite{Ren2019}. In particular, the architecture includes a Recurrent Neural Network (RNN) which is designed to utilize the sequential characteristic of the survival curve. It evaluates the hazard rate predictions at the discrete output time steps before and at the event time for each individual. Furthermore, the index $l \in \{ 1, \ldots, L\}$ of the current time step is an additional input feature to the encoder. Since the model predicts hazard rates $h_l$ for each discrete time point $t_l$, the survival curve prediction for individual $i$ is obtained by
\begin{equation}
    \hat{S}(t_l | x_i)=\prod_{j=1}^l (1-h_j).
\end{equation}
\subsubsection{Kamran}
The survival prediction model of \kamran{} \cite{Kamran2021} utilizes the architecture of \drsa{} but implements a novel loss function that emphasized proper model calibration. The first term, \lrps{}, uses the \textbf{R}ank \textbf{P}robability \textbf{S}core. Uncensored individuals contribute with a mean-squared-error loss that compares the individual's estimated survival function to a prediction that drops from one to zero at the discretized event time $z_i$ at the discrete output time point with index $l_i$. Moreover, censored individuals $(d_i = 0)$ only contribute up to the point of censoring $z_i$ and are compared to a prediction of one:
\begin{equation}\small %kamran LRPS
\label{eq:kamran:lrps}
\begin{split}
    \mathcal{L}_{\text {RPS}} = \sum_{i=1}^{n} \Bigg[ d_{i} \cdot & \sum_{l=1}^{L}\left(\hat{S}\left(t_l \mid \mathbf{x}_{i}\right)-\mathbbm{1}_{l<l_i}\right)^{2} + \\
    \left(1 - d_{i}\right) \cdot & \sum_{l=1}^{l_i}\left(\hat{S}\left(t_l \mid \mathbf{x}_{i}\right)-1\right)^{2} \Bigg].
\end{split}
\end{equation}
Here, $\hat{S}(t_l | \mathbf{x}_i)$ denotes the estimated survival curve at time $t_l$ with the individual $i$'s feature vector $\mathbf{x}_i$, and the corresponding index $l_i$ of the discrete event time $z_i$. The second loss term \lkernel{} emphasizes proper discrimination of the model by penalizing the wrong ordering of two \emph{uncensored} individuals
\begin{equation}\small
\label{eq:kamran:lkernel}
\begin{split}
    \mathcal{L}_{\text {kernel}} = \sum_{i=1}^n \sum_{j=1}^n A_{i, j} \cdot \exp \left(-\frac{1}{\sigma}\left(\hat{S}\left(z_{i} \mid \mathbf{x}_{j}\right)-\hat{S}\left(z_{i} \mid \mathbf{x}_{i}\right)\right)\right)
\end{split}
\end{equation}
where every non-zero entry of $A_{i, j}=\mathbbm{1}_{\left(i \neq j, d_{i}=d_{j}=1, z_{i}<z_{j}\right)}$ corresponds to one comparison of two non-censored individuals $i$ and $j$. Since $z_i < z_j$, the prediction $\hat{S}(z_i | x_i)$ should be smaller than $\hat{S}(z_i | x_j)$ to minimize this loss for all comparisons. Note that large values for $\sigma$ will emphasize the spreading of predictions for uncensored pairs. Lastly, the two loss parts are combined with a hyperparameter $\lambda$ for weighting to
\begin{equation}
\label{eq:kamran:loss}
\mathcal{L}_{\text {Kamran}} = \mathcal{L}_{\text {RPS}} + \lambda \mathcal{L}_{\text {kernel}}.
\end{equation}
\kamran{} uses a discrete output node per month and \lrps{} is evaluated on the survival curve $\hat{S}(t | x_i)$ directly, instead of the hazard rates $h_l$ as in \drsa{}.
\subsubsection{\dcs{}}
\label{models:ours}
\begin{figure}[!b]
\centerline{\includegraphics[width=\linewidth]{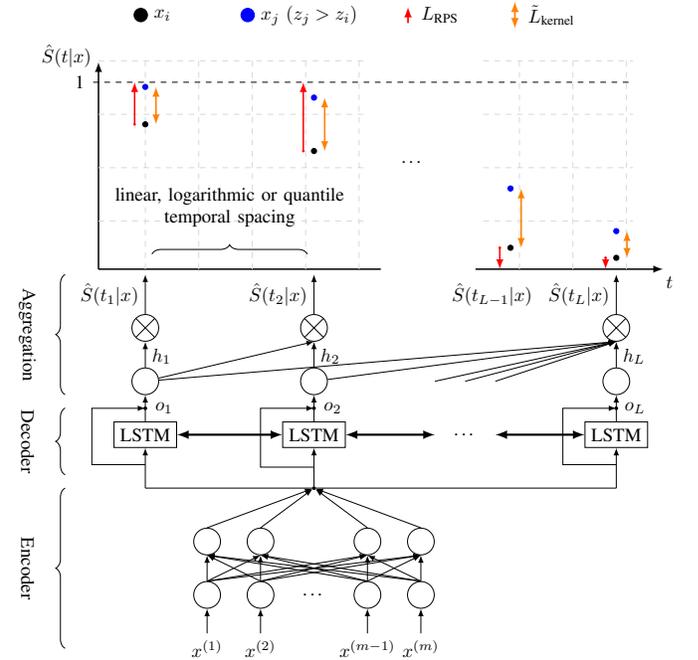}}
\caption{Visualization of the DCS network architecture from individual features $\mathbf{x}$ to the output of a survival curve prediction $\hat{S}(t|\mathbf{x})$. The red arrows indicate the \lrps{} loss, which minimizes the distance of the prediction to 1, before the event, and 0, after the event. The novel \dcs{} \lkernelours{} loss maximizes the distance between $x_i$ and $x_j$, as shown in orange.}
\label{fig:ours:architecture}
\end{figure}

Our new \dcs{} model can be divided into an encoder, decoder, and aggregation part and predicts a survival curve $\hat{S}(t|\mathbf{x})$ given a feature vector $\mathbf{x}$, as illustrated in \figureref{fig:ours:architecture}.
\textit{Model Architecture} - The basic architecture of \dcs{} is an extension and modification of \drsa{} that we will highlight in this section. 
\textbf{Encoder}: In contrast to DRSA, the encoder network does not append the index $l$ to the encoded feature vector $\mathbf{x}$ since it did not improve our results. This leads to a simplified encoding that is the same for every predicted time-step. Compared to DRSA, we also allow multiple fully connected encoder layers.
\textbf{Decoder}: The LSTM structure is extended to allow a skip connection that concatenates the original encoder output to the LSTM output for every timestep. Additionally, we introduce a bidirectional LSTM to maximize the usage of temporal information forwards and backwards in time.
\textbf{Aggregation}: To combine the LSTM's output for each timestep $t_l$ to a scalar hazard rate $h_l$, a fully connected network of one or more layers is used instead of only a single layer.
\textit{Loss Function} - \dcs{} features a modified kernel loss term \lkernel{} to optimize the use of both censored and uncensored patient data for better discriminative performance. The original kernel loss \eqref{eq:kamran:lkernel} is extended to not only include event-to-event (EE), but also event-to-censoring (EC) comparisons of two individuals $i$ and $j$ where the individual $j$ with the latter event time ($z_i < z_j$) is censored ($d_j = 0$). This modification increases the number of comparable training data pairs to boost discriminative performance. In more detail, we relax the condition ($d_{i}=d_{j}=1, z_{i}<z_{j}$) of the masking matrix $A$ in \eqref{eq:kamran:lkernel}, that only includes EE comparisons, to our new condition ($d_{i}=1, z_{i}<z_{j}$) that adds EC comparisons in the novel masking matrix $B$, resulting in the following kernel loss:
\begin{equation}\small
\label{eq:ours:lkernel}
\begin{split}
    \mathcal{\tilde{L}}_{\text {kernel}} = \sum_{i=1}^n \sum_{j=1}^n B_{i, j} \cdot \exp \left[-\frac{1}{\sigma}\left(\hat{S}\left(z_{i} \mid \mathbf{x}_{j}\right)-\hat{S}\left(z_{i} \mid \mathbf{x}_{i}\right)\right)\right],
\end{split}
\end{equation}
where $B_{i, j}=\mathbbm{1}_{\left(i \neq j, d_{i}=1, z_{i}<z_{j}\right)}$. This increases the maximum number of comparisons for e.g. the \support{} dataset from \num{1.81e7} to \num{3.41e7}, a factor of $F=1.9$. The factor $F$ depends on the censoring rate and the censoring distribution of each dataset as shown in \tableref{tbl:loss:comparisons}. Moreover, this work analyzes the relation between the censoring rate and the number of comparisons for a dataset in detail in the supplemental material. The second \dcs{} loss term is identical to the RPS calibration loss of \kamran{} \eqref{eq:kamran:lrps}.  We introduced a normalization step for the two independent losses to achieve better orthogonalization of the hyperparameters batch size and $\lambda$. Without this normalization, the batch size would influence the choice of $\lambda$ since the weighting of the loss terms depends on the number of individuals $n$ and the number of output nodes $L$ that might be included in the hyperparameter search space.  While \lrps{} $\in \mathcal{R}^{n \times L}$ is normalized by its width $L$ and height $n$, \lkernel{} $\in \mathcal{R}^{n \times n}$ is a sparse matrix that is normalized by the number of actual comparisons $n_{comp} = |B|$. This normalization step leads to our final objective function:
\begin{equation}\small
\begin{split}
    \mathcal{L}_{ours} = \frac{1}{nL} \mathcal{L}_{RPS} +  \frac{\lambda}{n_{comp}} \mathcal{\tilde{L}}_{kernel}
\end{split}
\end{equation}
\textit{Temporal Spacing of Predictions} - \kamran{} and \drsa{} use equidistant temporal spacing (\emph{linear spacing}) of survival predictions, such that every consecutive inference predicts the survival of a patient for e.g. another month. While equidistant temporal predictions are a natural choice, they might not be optimal for model learning given the training data distribution. To obtain a model with high-quality predictions for earlier as well as later time-points, we propose two alternative temporal output node spacings to get a more uniform distribution of events per time-point, as outlined in \figureref{fig:spacing:support}. Histograms for the other datasets in this work can be found in the supplemental material. As a medical motivation, one might argue that the difference between surviving $100$ and $200$ days should be more important than the difference between $1100$ and $1200$ days, thus motivating a \emph{logarithmic spacing} of predictions. To enforce a strictly uniform distribution of training events per time-point, one can ensure that at each time-point an equal amount of individuals either experience the event or are censored, which we call \emph{quantile spacing}.
\begin{table}[bp] % no. comparison
\begin{center}
\caption{Number of comparisons $|A|$ and $|B|$ for each analyzed dataset that corresponds to the number of non-zero entries in $A$ and $B$ respectively. Also included is the number of patients, censoring rate and the factor of more comparisons $F = |B|/|A|$.}
\resizebox{\linewidth}{!}{
\begin{tabular}{lrrrrr}
 & patients & censoring rate & $|A|$ & $|B|$ & $F$ \\ \midrule
\support{} & 8873 & \SI{32}{\percent} & \num{1.81e7} & \num{3.41e7} & 1.9 \\
\metabric{} & 1904 & \SI{42}{\percent} & \num{6.08e5} & \num{1.25e6} & 2.0 \\
\flchain{} & 7874 & \SI{72}{\percent} & \num{2.35e6} & \num{1.34e7} & 5.7
\end{tabular}
}
\label{tbl:loss:comparisons}
\end{center}
\end{table}
\subsection{Metrics}
\subsubsection{Discrimination}
Discrimination quantifies if a model predicts the correct order of events, it does not scrutinize if the predicted event occurs at the correct time. An example for the usage of discrimination is an organ transplant waiting list, where the patient with the best survival estimate after transplantation is selected as the most suitable candidate. In this work we measure discrimination with the \cindextd{} (\cindextdabb{}) \cite{Antolini2005} and the \cdauc{} (\cdaucabb{}) \cite{Uno2007}.
 
\textit{\cindextd{}} - The \cindexabb{} \cite{Harrell1982} is the most commonly used metric for discriminative performance of survival models. It is a generalization of the area under the ROC curve (AUC) \cite{Antolini2005} that evaluates the correct ordering of predictions compared to actual event times. Several models we evaluate allow crossing survival curve predictions (i.e.\ the order of individuals can change over time). To properly evaluate these time-dependent changes, we use the \cindextd{} (\cindextdabb{}) \cite{Antolini2005}, which evaluates the \cindexabb{} at the event times.
\begin{equation}\small
\label{eq:cindex}
\begin{split}
    C^{td} = P(\hat{S}(z_i|x_i) < \hat{S}(z_i|x_j)\ |\  z_i<z_j,\,d_i=1)
\end{split}
\end{equation}
\begin{figure}[b] % figure grids support
    \centering
    \includegraphics[width=\linewidth]{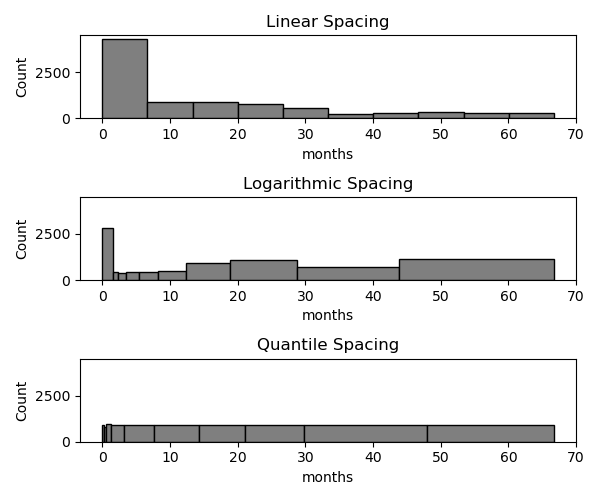}
    \label{fig:output_grids:support}
    \caption{Histograms of the number of events per temporal prediction on \support{}. Ten predictions in time are made with linear, logarithmic and quantile temporal spacing.}
    \label{fig:spacing:support}
\end{figure}
\textit{\cdauc{} (\cdaucabb{})} - To measure how well a model can distinguish individuals that experience an event before or at time $z_i \leq t$ from those that did not experience an event yet ($z_i > t$), a time dependent receiver operating characteristic curve (ROC) and the corresponding area under the curve can be defined as \cite{Blanche2019}:
\begin{equation}\small
    \label{eq:AUC}
\begin{split}
    %AUC(t) = \frac{\sum_{i=1}^n \sum_{j=1}^n \mathbbm{1}_{z_i \leq t, z_j > t, \hat{S}(t|x_i) \geq \hat{S}(t|x_j)}}{\sum_{i=1}^n \mathbbm{1}_{z_i \leq t} \sum_{j=1}^n \mathbbm{1}_{z_j > t}}
    AUC(t) = P ( \hat{S}(t|x_i) < \hat{S}(t|x_j )\ |\ z_i \leq t,\,z_j > t)
\end{split}
\end{equation}
To generate a scalar metric for the whole time span of the dataset, the $AUC(t)$ is weighted (by the KM estimate $\tilde{S}(t)$) and integrated by the inverse probability of censoring:
\begin{equation}\small
\label{eq:CDAUC}
\begin{split}
    CDAUC := \frac{1}{\tilde{S}(\tau_1) - \tilde{S}(\tau_2)} \int_{\tau_1}^{\tau_2} AUC(t)\,d\tilde{S}(t).
\end{split}
\end{equation}
As opposed to the \cindexabb{}, the \cdaucabb{} is a proper scoring method. No prediction model has a higher discriminative value than the data generating process itself\cite{Blanche2019}. 
\subsubsection{Calibration}
A patient or medical practitioner might be interested if the prediction reflects the true underlying survival distribution. A relapse time of 2 or 20 years will make a difference for the selected therapy, independent of the ordering of other patients. We measure calibration with the \ddcabb{} \cite{Kamran2021}.
\textit{\ddc{} (\ddcabb{})} - The estimated survival for each individual at their event time $\hat{S}(z_i|\mathbf{x}_i)$ is mapped into $10$ equally sized bins $\mathcal{B}$ in the unit interval as shown in \cite{Haider2018}. For \ddcabb{}, the KL-Divergence \cite{kullback1951} is calculated between the relative frequency $P$ of the observed intervals compared to the uniform distribution $Q(b) = 0.1$:
\begin{equation}\small
\begin{split}
\label{eq:DDC}
    DDC(P \| Q)=\sum_{b \in \mathcal{B}} P(b) \log \left(\frac{P(b)}{Q(b)}\right)
\end{split}
\end{equation}
\section{Experiments}
\label{sec:experiments}
\subsection{Datasets}
We evaluated the presented models on the following three public medical EHR datasets to analyze the discrimination and calibration performance. The basic and relevant properties of those datasets are depicted in \tableref{tab:datasets}.\\

\begin{table}[b] 
\begin{center}
\caption{Number of patients, number of features, censoring rate, percent of missing data, median survival time and ratio of covariates that violate the proportional hazards assumption (non-prop.) for each dataset.}
% \resizebox{\linewidth}{!}{%
\begin{tabular}{lrrr}
 & \support{} & \metabric{} & \flchain{} \\
\midrule
 patients & 8873 & 1904 & 7874 \\
 features & 14 & 9 & 8 \\
 censoring rate & \SI{32}{\percent} & \SI{42}{\percent} & \SI{72}{\percent} \\
 missing data & \SI{0.1}{\percent} & \SI{0}{\percent} & \SI{1.7}{\percent} \\
median surv. & \SI{57}{days} & \SI{85}{months} & \SI{71}{months} \\
non-prop. & \SI{79}{\percent} & \SI{56}{\percent} & \SI{25}{\percent}
\end{tabular}
% }
\label{tab:datasets}
\end{center}
\end{table}

\textit{\support{} \cite{Knaus1995}} - The Study to Understand Prognoses and Preferences for Outcomes and Risks of Treatments provides information about 8873 seriously ill hospitalized adults with 14 features (demographic, observational, and lab data). The endpoint is the time of death of a patient.

\textit{\metabric{} \cite{curtis2012}} - This dataset from the Molecular Taxonomy of Breast Cancer International Consortium is the smallest analyzed dataset of this work. 1904 patients with 9 features (demographic, molecular drivers and therapies) are given to predict long-term clinical outcome of breast cancer.

\textit{\flchain{} \cite{Therneau2014}} - This study analyzes the impact of free light chain and creatinine blood serum levels. They also include basic demographic information in their analysis to predict patient mortality.

\subsection{Implementation}
% \footnote{\href{https://github.com/CamDavidsonPilon/lifelines/}{https://github.com/CamDavidsonPilon/lifelines/}}
%\footnote{\href{https://github.com/sebp/scikit-survival}{https://github.com/sebp/scikit-survival}}
%\footnote{\href{https://github.com/scikit-learn/scikit-learn}{https://github.com/scikit-learn/scikit-learn}}
%\footnote{\href{https://github.com/havakv/pycox}{https://github.com/havakv/pycox}}
In our experiments we use the \coxph{} model from \verb~lifelines~\cite{lifelines}, \deepsurv{} and \coxtime{} from \verb~pycox~ \cite{Kvamme2019}. The \drsa{}, \kamran{} and our \dcs{} models were self-implemented in Tensorflow 2.5.0 and Python 3.8.7. The implementation uses external implementations of the \cindextdabb{} (\verb~pycox~) and \cdaucabb{}  (\verb~scikit-survival~ \cite{sksurv}). The \ddc{} was self-implemented. For automated hyperparameter tuning we used \verb~scikit-learn~ \cite{scikit-learn} wrappers for all models, preprocessing workflows, and datasets. Our code with all datasets, baseline models, pipelines, hyperparameter tuning, and metrics can be found in our repository at \href{https://github.com/imsb-uke/dcsurv}{https://github.com/imsb-uke/dcsurv}.

\subsection{Data Processing}

The dataset features are standardized and imputed, where required. Numerical features are standardized to a mean of zero and a standard deviation of one. To impute missing values, we use \emph{median imputation} for numerical and \emph{most frequent} imputation for categorical features. Categorical features are one-hot encoded. For the \support{} dataset, zeros in some continuous features like heart rate or respiration rate were also treated as missing values and imputed accordingly.
To evaluate the continuous and discrete time models as fair as possible, all estimations are projected on the same time-steps. We chose the training set event and censoring times for evaluation. For this purpose, all model predictions were interpolated linearly between their temporal output node time points.
\subsection{Hyperparameter Tuning}
For hyperparameter tuning and performance evaluation we split the data into 80\% training and 20\% test data stratified by the event indicator. The test data was only used during the final evaluation of the model performances. A detailed description of the hyperparameter tuning and the best parameters for each model can be found in the supplements.
\section{Results}
\label{sec:results}
We compare the discrimination and calibration performance of \coxph{}, \deepsurv{}, \coxtime{}, \drsa{}, \kamran{}, and three variants of \dcs{} with linear, logarithmic, and quantile output node spacing on the \support{}, \metabric{}, and \flchain{} test data. To obtain variance estimations for the model performance, we use 10-fold bootstrapping and report the 'mean $\pm$ standard deviation'. In the evaluation we focus on three scenarios, the overall best model per metric, the performance gains achieved by \dcs{}'s novel kernel loss and our novel \textbf{lin}ear, \textbf{log}arithmic and \textbf{quant}ile spacing approaches. Overall, \dcs{} models reach best discriminative performance for both the \cindextdabb{} (\tableref{table:quant_results:cindex}) and the \cdaucabb{} (\tableref{table:quant_results:cdauc}). More specifically, the \dcs{} model with quantile output node spacing (DCS-quant) displays top performance in four out of six comparisons. DCS-quant shows the best \cindextdabb{} on the \support{} ($0.628$) and \metabric{} ($0.698$) datasets, while DCS-linear reaches first place on the \flchain{} dataset ($0.803$) (\tableref{table:quant_results:cindex}). Regarding the \cdaucabb{}, DCS-quant exhibits top performance on the \metabric{} ($0.773$) and \flchain{} ($0.832$) datasets, while the DCS-log model is best on the \support{} data ($0.658$) (\tableref{table:quant_results:cdauc}). These results suggest that both, the novel kernel loss \eqref{eq:ours:lkernel} and new output spacing, increase discriminative performance as measured with the \cindextdabb{} and \cdaucabb{}, robustly outperforming all competing models. To specifically understand the influence of the modified kernel loss \eqref{eq:ours:lkernel} on discriminative performance, we compare DCS-linear to \kamran{}. For all three datasets, DCS-linear outperforms \kamran{} in the \cindextdabb{} (\support{} $0.628$ vs. $0.610$, \metabric{} $0.694$ vs. $0.668$, \flchain{} $0.803$ vs. $0.786$). Similarly, DCS-linear surpasses \kamran{} in the \cdaucabb{} on \metabric{} ($0.730$ vs. $0.727$) and \flchain{} ($0.813$ vs. $0.807$), while it is slightly worse on the \support{} ($0.641$ vs. $0.652$) data. Lastly, it is interesting to observe that the continuous time models \coxph{}, \deepsurv{}, and \coxtime{} outperform the discrete \drsa{} and \kamran{} models in the \cindextdabb{}. Continuous time models reach the best calibration as measured by \ddcabb{}, while DCS-quant reaches the best calibration performance of all discrete time models (\tableref{table:quant_results:ddc}). For instance, \coxtime{} features a slightly lower (better) \ddcabb{} as compared to DCS-quant accross all datasets (\support{} $0.066$ vs. $0.007$, \metabric{} $0.027$ vs. $0.009$, \flchain{} $0.021$ vs. $0.006$). Among the discrete time models, our DCS-quant shows superior performance compared to \drsa{} on all datasets (\support{} $0.066$ vs. $0.193$, \metabric{} $0.027$ vs. $0.296$, \flchain{} $0.021$ vs. $0.050$) and to \kamran{} on two out of three datasets (\support{} ($0.066$ vs. $0.139$, \metabric{} $0.027$ vs. $0.065$, \flchain{} $0.021$ vs. $0.006$). These results provide strong evidence that (quantile) output node spacing can boost both, calibration and discrimination performance of survival models.
\begin{table}[htb] % c-index
\begin{center}
\caption{Test set results for \cindextdabb{} ($\uparrow$).}
\vspace*{-1em}
\resizebox{\linewidth}{!}{%
\begin{tabular}{llrrr}
& &                     \support &                    \metabric &                     \flchain \\
\midrule
\multirow{3}{*}{\rotatebox{90}{cont.}} & CoxPH              &           $0.594 \pm 0.009$ &           $0.638 \pm 0.020$ &           $0.798 \pm 0.007$ \\
& DeepSurv           &           $0.604 \pm 0.012$ &           $0.679 \pm 0.014$ &           $0.795 \pm 0.014$ \\
& CoxTime            &           $0.612 \pm 0.007$ &           $0.675 \pm 0.019$ &           $0.790 \pm 0.010$ \\
\midrule
\multirow{2}{*}{\rotatebox{90}{disc.}} & \drsa{}               &           $0.598 \pm 0.006$ &           $0.661 \pm 0.019$ &           $0.792 \pm 0.018$ \\
& \kamran{}             &           $0.610 \pm 0.006$ &           $0.668 \pm 0.023$ &           $0.786 \pm 0.013$ \\
\midrule
\multirow{3}{*}{\rotatebox{90}{ours}} & DCS-linear   &           $0.623 \pm 0.009$ &           $0.694 \pm 0.018$ &  $\mathbf{0.803 \pm 0.011}$ \\
& DCS-log      &           $0.623 \pm 0.009$ &           $0.674 \pm 0.018$ &           $0.792 \pm 0.008$ \\
& DCS-quant &  $\mathbf{0.628 \pm 0.009}$ &  $\mathbf{0.698 \pm 0.019}$ &           $0.794 \pm 0.015$ \\
\end{tabular}
}
\label{table:quant_results:cindex}
\end{center}
\end{table}
\vspace*{-1em}
\begin{table}[htb] % cdauc
\begin{center}
\caption{Test set results for \cdaucabb{} ($\uparrow$).}
\vspace*{-1em}
\resizebox{\linewidth}{!}{%
\begin{tabular}{llrrr}
& &                     \support &                    \metabric &                     \flchain \\
\midrule
\multirow{3}{*}{\rotatebox{90}{cont.}} & CoxPH              &           $0.619 \pm 0.013$ &           $0.686 \pm 0.028$ &           $0.797 \pm 0.019$ \\
& DeepSurv           &           $0.634 \pm 0.016$ &           $0.700 \pm 0.032$ &           $0.800 \pm 0.016$ \\
& CoxTime            &           $0.647 \pm 0.010$ &           $0.747 \pm 0.020$ &           $0.799 \pm 0.012$ \\
\midrule
\multirow{2}{*}{\rotatebox{90}{disc.}} & \drsa{}               &           $0.613 \pm 0.008$ &           $0.685 \pm 0.029$ &           $0.814 \pm 0.014$ \\
& \kamran{}             &           $0.652 \pm 0.018$ &           $0.727 \pm 0.024$ &           $0.807 \pm 0.018$ \\
\midrule
\multirow{3}{*}{\rotatebox{90}{ours}} & DCS-linear   &           $0.641 \pm 0.012$ &           $0.730 \pm 0.021$ &           $0.813 \pm 0.017$ \\
& DCS-log      &  $\mathbf{0.658 \pm 0.011}$ &           $0.716 \pm 0.028$ &           $0.824 \pm 0.015$ \\
& DCS-quant &           $0.657 \pm 0.012$ &  $\mathbf{0.773 \pm 0.023}$ &  $\mathbf{0.832 \pm 0.017}$ \\
\end{tabular}
}
\label{table:quant_results:cdauc}
\end{center}
\end{table}
\vspace*{-1em}
\begin{table}[htb] % ddc
\begin{center}
\caption{Test set results for \ddcabb{} ($\downarrow$).}
\vspace*{-1em}
\resizebox{\linewidth}{!}{%
\begin{tabular}{llrrr}
& &                     \support &                    \metabric &                     \flchain \\

\midrule
\multirow{3}{*}{\rotatebox{90}{cont.}} & CoxPH              &           $0.009 \pm 0.003$ &           $0.021 \pm 0.007$ &  $\mathbf{0.001 \pm 0.000}$ \\
& DeepSurv           &           $0.010 \pm 0.003$ &           $0.017 \pm 0.006$ &           $0.003 \pm 0.002$ \\
& CoxTime            &  $\mathbf{0.007 \pm 0.002}$ &  $\mathbf{0.009 \pm 0.004}$ &           $0.006 \pm 0.001$ \\
\midrule
\multirow{2}{*}{\rotatebox{90}{disc.}} & \drsa{}               &           $0.193 \pm 0.009$ &           $0.296 \pm 0.024$ &           $0.050 \pm 0.004$ \\
& \kamran{}             &           $0.139 \pm 0.006$ &           $0.065 \pm 0.013$ &           $0.006 \pm 0.002$ \\
\midrule
\multirow{3}{*}{\rotatebox{90}{ours}} & DCS-linear   &           $0.141 \pm 0.013$ &           $0.138 \pm 0.010$ &           $0.012 \pm 0.002$ \\
& DCS-log      &           $0.055 \pm 0.005$ &           $0.071 \pm 0.006$ &           $0.035 \pm 0.004$ \\
& DCS-quant &           $0.066 \pm 0.006$ &           $0.027 \pm 0.008$ &           $0.021 \pm 0.004$ \\
\end{tabular}
}
\label{table:quant_results:ddc}
\end{center}
\end{table}
\section{Conclusion}
In this work we present a novel DL-based survival model, \dcs{}, with state-of-the-art discrimination and good calibration. Two novel features that are introduced in \dcs{} seem to be primarily responsible for this performance increase. First, we introduce an extension of the \kamran{} \lkernel{} loss \cite{Kamran2021}, which uses event-to-event and event-to-censoring pairs during training to boost discrimination. We show that our \lkernelours{} increases the number of comparisons for the analyzed datasets by a factor of approximately $2$ to $6$ as compared to the original implementation by \kamran{}. These additional comparisons improve the discriminative performance regarding \cindextdabb{} as well as \cdaucabb{} on the three analyzed datasets. Second, we introduce three temporal output node spacings (linear, logarithmic, and quantile spacing) to increase the model discrimination and calibration. Overall, the quantile approach (DCS-quant) provides the best discriminative performance on the three datasets we investigated. DCS-quant also reaches the best calibration for discrete time models (\drsa{} and \kamran{}) for two of three datasets. While DCS-quant shows best calibration for discrete time models, it is slightly worse than the calibration of the continuous time models (\coxph{}, \deepsurv{}, and \coxtime{}). A reason for the good calibration of continuous time models, at the cost of inferior discrimination, might be that the time-dependent baseline hazard already defines a reasonable population wide survival estimate. This could suggest that continuous models might suffer from bad calibration on datasets with time-dependent underlying baseline hazards. Our results also indicate that the \cindextdabb{} might not be the best metric to evaluate discriminative performance in survival prediction. Whereas the \cindextdabb{} for the \flchain{} dataset yields very similar values for all models inspected, the \cdaucabb{} shows clear differences in model performance. In the end, \dcs{} provides researchers with a model that reaches best-of-breed discrimination and good calibration and might pave the way towards future clinical application of deep-learning-based survival prediction.

\section*{Acknowledgements}
PF and AE were supported by the KFO 306 and LFF-FV 78 grants. ED received funding from the SFB 1192 project B8. FW was supported by the UKE M3I grant and SB by the UKE R3 reduction of animal testing grant. KK received funding from the UKE deanery.

\bibliographystyle{IEEEtran}
\bibliography{library}

\begin{thebibliography}{10}
\providecommand{\url}[1]{#1}
\csname url@samestyle\endcsname
\providecommand{\newblock}{\relax}
\providecommand{\bibinfo}[2]{#2}
\providecommand{\BIBentrySTDinterwordspacing}{\spaceskip=0pt\relax}
\providecommand{\BIBentryALTinterwordstretchfactor}{4}
\providecommand{\BIBentryALTinterwordspacing}{\spaceskip=\fontdimen2\font plus
\BIBentryALTinterwordstretchfactor\fontdimen3\font minus
  \fontdimen4\font\relax}
\providecommand{\BIBforeignlanguage}[2]{{%
\expandafter\ifx\csname l@#1\endcsname\relax
\typeout{** WARNING: IEEEtran.bst: No hyphenation pattern has been}%
\typeout{** loaded for the language `#1'. Using the pattern for}%
\typeout{** the default language instead.}%
\else
\language=\csname l@#1\endcsname
\fi
#2}}
\providecommand{\BIBdecl}{\relax}
\BIBdecl

\bibitem{Klein2003}
J.~P. Klein and M.~L. Moeschberger, \emph{Survival analysis: techniques for
  censored and truncated data}.\hskip 1em plus 0.5em minus 0.4em\relax
  Springer, 2003, vol.~2.

\bibitem{Kaplan1958}
E.~L. Kaplan and P.~Meier, ``Nonparametric estimation from incomplete
  observations,'' \emph{Journal of the American Statistical Association},
  vol.~53, no. 282, pp. 457--481, 1958.

\bibitem{Cox1972}
D.~R. Cox, ``Regression models and life-tables,'' \emph{Journal of the Royal
  Statistical Society: Series B (Methodological)}, vol.~34, no.~2, pp.
  187--202, 1972.

\bibitem{Meng2021}
Z.~Meng \emph{et~al.}, ``A multi-task kernel learning algorithm for survival
  analysis,'' in \emph{Pacific-Asia Conference on Knowledge Discovery and Data
  Mining}.\hskip 1em plus 0.5em minus 0.4em\relax Springer, 2021, pp. 298--311.

\bibitem{Haider2018}
H.~Haider \emph{et~al.}, ``Effective ways to build and evaluate individual
  survival distributions.'' \emph{J. Mach. Learn. Res.}, vol.~21, no.~85, pp.
  1--63, 2020.

\bibitem{Katzman2018}
\BIBentryALTinterwordspacing
J.~L. Katzman \emph{et~al.}, ``{DeepSurv: Personalized treatment recommender
  system using a Cox proportional hazards deep neural network},'' \emph{BMC
  Medical Research Methodology}, vol.~18, no.~1, pp. 1--12, feb 2018. [Online].
  Available: \url{https://link.springer.com/articles/10.1186/s12874-018-0482-1}
\BIBentrySTDinterwordspacing

\bibitem{Kvamme2019}
\BIBentryALTinterwordspacing
H.~Kvamme \emph{et~al.}, ``{Time-to-event prediction with neural networks and
  Cox Regression},'' \emph{Journal of Machine Learning Research}, vol.~20, pp.
  1--30, 2019. [Online]. Available:
  \url{http://jmlr.org/papers/v20/18-424.html.}
\BIBentrySTDinterwordspacing

\bibitem{Ren2019}
\BIBentryALTinterwordspacing
K.~Ren \emph{et~al.}, ``{Deep recurrent survival analysis},'' \emph{33rd AAAI
  Conference on Artificial Intelligence, AAAI 2019, 31st Innovative
  Applications of Artificial Intelligence Conference, IAAI 2019 and the 9th
  AAAI Symposium on Educational Advances in Artificial Intelligence, EAAI
  2019}, pp. 4798--4805, sep 2019. [Online]. Available:
  \url{http://arxiv.org/abs/1809.02403}
\BIBentrySTDinterwordspacing

\bibitem{Kamran2021}
\BIBentryALTinterwordspacing
F.~Kamran and J.~Wiens, ``{Estimating Calibrated Individualized Survival Curves
  with Deep Learning},'' \emph{The Thirty-Fifth AAAI Conference on Artificial
  Intelligence}, vol.~35, no.~1, pp. 240--248, 2021. [Online]. Available:
  \url{https://ojs.aaai.org/index.php/AAAI/article/view/16098}
\BIBentrySTDinterwordspacing

\bibitem{park2015}
S.~Park and D.~J. Hendry, ``Reassessing schoenfeld residual tests of
  proportional hazards in political science event history analyses,''
  \emph{American Journal of Political Science}, vol.~59, no.~4, pp. 1072--1087,
  2015.

\bibitem{Antolini2005}
\BIBentryALTinterwordspacing
L.~Antolini \emph{et~al.}, ``{A time-dependent discrimination index for
  survival data},'' \emph{Statistics in Medicine}, vol.~24, no.~24, pp.
  3927--3944, dec 2005. [Online]. Available:
  \url{http://doi.wiley.com/10.1002/sim.2427}
\BIBentrySTDinterwordspacing

\bibitem{Uno2007}
\BIBentryALTinterwordspacing
H.~Uno \emph{et~al.}, ``{Evaluating prediction rules for t-year survivors with
  censored regression models},'' \emph{Journal of the American Statistical
  Association}, vol. 102, no. 478, pp. 527--537, jun 2007. [Online]. Available:
  \url{https://www.tandfonline.com/doi/abs/10.1198/016214507000000149}
\BIBentrySTDinterwordspacing

\bibitem{Harrell1982}
\BIBentryALTinterwordspacing
F.~E. Harrell \emph{et~al.}, ``{Evaluating the Yield of Medical Tests},''
  \emph{JAMA}, vol. 247, no.~18, pp. 2543--2546, may 1982. [Online]. Available:
  \url{https://jamanetwork.com/journals/jama/fullarticle/372568}
\BIBentrySTDinterwordspacing

\bibitem{Blanche2019}
P.~Blanche \emph{et~al.}, ``{The c-index is not proper for the evaluation of
  t-year predicted risks},'' \emph{Biostatistics}, vol.~20, no.~2, pp.
  347--357, 2019.

\bibitem{kullback1951}
\BIBentryALTinterwordspacing
S.~Kullback and R.~A. Leibler, ``{On Information and Sufficiency},''
  \emph{https://doi.org/10.1214/aoms/1177729694}, vol.~22, no.~1, pp. 79--86,
  mar 1951. [Online]. Available:
  \url{https://projecteuclid.org/journals/annals-of-mathematical-statistics/volume-22/issue-1/On-Information-and-Sufficiency/10.1214/aoms/1177729694.full}
\BIBentrySTDinterwordspacing

\bibitem{Knaus1995}
W.~A. Knaus \emph{et~al.}, ``{The SUPPORT prognostic model. Objective estimates
  of survival for seriously ill hospitalized adults},'' \emph{Annals of
  Internal Medicine}, vol. 122, no.~3, pp. 191--203, 1995.

\bibitem{curtis2012}
C.~Curtis \emph{et~al.}, ``The genomic and transcriptomic architecture of 2,000
  breast tumours reveals novel subgroups,'' \emph{Nature}, vol. 486, no. 7403,
  pp. 346--352, 2012.

\bibitem{Therneau2014}
\BIBentryALTinterwordspacing
T.~M. Therneau, \emph{A Package for Survival Analysis in R}, 2022, r package
  version 3.3-1. [Online]. Available:
  \url{https://CRAN.R-project.org/package=survival}
\BIBentrySTDinterwordspacing

\bibitem{lifelines}
\BIBentryALTinterwordspacing
C.~Davidson-Pilon \emph{et~al.}, \emph{CamDavidsonPilon/lifelines: v0.25.7},
  Dec. 2020. [Online]. Available: \url{https://doi.org/10.5281/zenodo.4313838}
\BIBentrySTDinterwordspacing

\bibitem{sksurv}
\BIBentryALTinterwordspacing
S.~P{\"o}lsterl, ``scikit-survival: A library for time-to-event analysis built
  on top of scikit-learn,'' \emph{Journal of Machine Learning Research},
  vol.~21, no. 212, pp. 1--6, 2020. [Online]. Available:
  \url{http://jmlr.org/papers/v21/20-729.html}
\BIBentrySTDinterwordspacing

\bibitem{scikit-learn}
F.~Pedregosa \emph{et~al.}, ``Scikit-learn: Machine learning in {P}ython,''
  \emph{Journal of Machine Learning Research}, vol.~12, pp. 2825--2830, 2011.

\end{thebibliography}

\end{document}

% End Crossreference between files

 % IEEE HEADER
 
% Header

\title{Deep Learning-Based Discrete Calibrated Survival Prediction\\ \Large{Supplemental Material}}
\author{\IEEEauthorblockN{Patrick Fuhlert, Anne Ernst, Esther Dietrich, Fabian Westhaeusser, Karin Kloiber, Stefan Bonn}
\IEEEauthorblockA{\textit{Institute of Medical Systems Biology, Center for Biomedical AI (bAIome), Center for Molecular Neurobiology (ZMNH)}\\
\textit{University Medical Center Hamburg-Eppendorf}, Hamburg, Germany}
}
{\maketitle}

\section{Estimating the number of comparisons for \cindextdabb{} and newly proposed loss}
\label{apx:no_comparisons}
Suppose a dataset $D$ with $n$ individuals and a censoring rate of $c \in [0, 1)$. When drawing two individuals $i$ and $j$ from $D$ at random, under the assumption that censoring is uniformly distributed throughout the observed event time, the chance of picking a pair $(i, j)$ that is comparable regarding the \cindextdabb{} \eqref{eq:cindex} or \lkernelours{} \eqref{eq:ours:lkernel} (EE or EC pair) is
\begin{equation}
\begin{split}
P(\text{comp}_\text{new}) = & P(d_i = 1, d_j = 1, z_i < z_j)\ +\\
&P(d_i = 1, d_j = 0, z_i < z_j)\\
\end{split}
\end{equation}
and can be estimated by using the censoring rate $c$ of randomly drawing a censored individual from the dataset. Under the assumption that, when picking a random pair $(i, j)$, the corresponding event times are equally distributed, the probability that the first event time $z_i$ is smaller than $z_j$ is assumed as $P(z_i < z_j) = 0.5$. Firstly, the relative frequency of drawing only EE pairs which corresponds to the number of comparisons in \eqref{eq:kamran:lkernel} can be estimated as

\begin{equation}
\begin{split}
\label{eq:pcomparable:old}
P(\text{comp}_{\text{old}}) = P(d_i = 1, d_j = 1, z_i < z_j) = (1-c)^2 / 2.
\end{split}
\end{equation}

Furthermore, we add the comparisons where the former event time is uncensored and the latter event time is censored as 

\begin{equation}
\begin{split}
P(d_i = 1, d_j = 0, z_i < z_j) &= c (1-c) / 2.
\label{eq:pcomparable:additional}
\end{split}
\end{equation}

Combining \eqref{eq:pcomparable:old} and \eqref{eq:pcomparable:additional} leads to the final estimation for the new number of comparisons
\begin{equation}
\begin{split}
\label{eq:pcomparable:new}
P(\text{comp}_\text{new}) = (1-c)^2 / 2 + c (1-c) / 2 = (1-c) / 2.
\end{split}
\end{equation}
Since the old $\mathcal{L}_{\text{kernel}}$ \eqref{eq:kamran:lkernel} only compared all uncensored pairs for $i$ and $j$ that led to $P(\text{comp}_{old}) = (1-c)^2$, the factor $F$ of more comparisons can be estimated from \eqref{eq:pcomparable:old} and \eqref{eq:pcomparable:new} as 
\begin{equation}
\begin{split}
    F_{\text{est}} = P(\text{comp}_\text{new}) / P(\text{comp}_{\text{old}}) = 1 / (1-c).
\label{eq:pcomparable:f}
\end{split}
\end{equation}
\begin{figure}[ht] % figure comparisons
\centering
    \subfigure[Number of normalized comparisons $n_{\text{comp}} / n^2$ and estimations (\ref{eq:pcomparable:old}, \ref{eq:pcomparable:new}) over the censoring rate. Old comparisons are depicted with "o" while new comparisons are shown with an "x".]
    {
        \includegraphics[width=\linewidth]{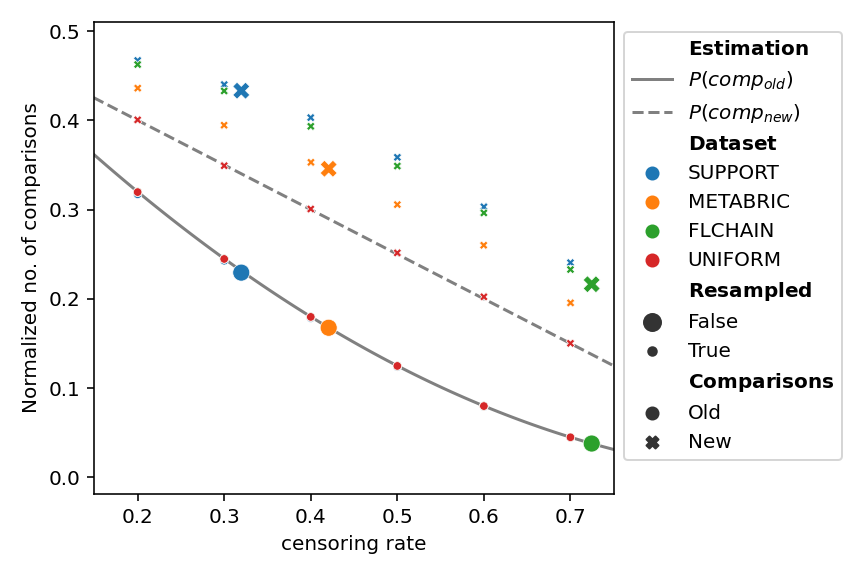}
        \label{fig:comparisons:relative}
    }
    \subfigure[Factor $F$ \eqref{eq:pcomparable:f} of more comparisons over the censoring rate.]
    {
        \includegraphics[width=\linewidth]{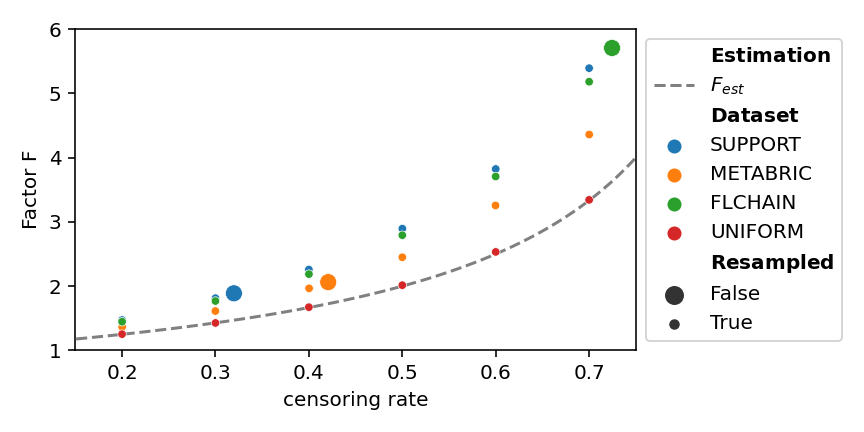}
        \label{fig:comparisons:normalized}
    }
    
\caption{Observed (markers) and estimated (lines) number of comparisons over the censoring rate for the analyzed datasets. We also include a synthetic dataset ('uniform') in red with a uniform censoring distribution. To generate additional datapoints, the datasets are resampled to match predefined censoring rates that are depicted with smaller markers.} 
\label{fig:comparisons}
\vspace*{-1em}
\end{figure}
This estimation can now be compared to the observed number of comparisons from \tableref{tbl:loss:comparisons} in \figureref{fig:comparisons}. It is shown that our estimation closely matches the number of old comparisons for all datasets as well as the number of new comparisons for the synthetically created new "uniform" dataset. It is shown that, for higher censoring rates, the factor $F$ suggests that our approach allows for significantly more comparisons on datasets with high censoring rates (e.g. \SI{70}{\percent} censoring leads to approximately $3$ to $6$ times more comparisons). Also notice that the observed factor $F$ is even higher than predicted for the real-world datasets which indicates that the assumption of a uniform censoring distribution is not met.
\section{Hyperparameter Tuning}
\label{apx:htuning}
This work utilizes a bayesian hyperparameter search with \verb~scikit-optimize~. The data was split into 80\% train and 20\% test data stratified by the event indicator. Afterwards, $5$-fold cross-validation on the training set was performed with $100$ iterations to find the best hyperparameters. The \cdaucabb{} was chosen as the optimization criterion. The best model was then retrained on the complete training dataset with the previously found best hyperparameters. We used a fixed batch size of $50$ and trained all models for $100$ epochs. To avoid overfitting during training, we added early stopping with a patience of $10$ epochs and included a \SI{20}{\percent} dropout rate.

The tunable hyperparameters of the \deepsurv{} and \coxtime{} models were the number of layers and nodes per layer of the networks. \coxph{} did not have any tunable hyperparameters. In \drsa{}, the loss weighting parameter $\alpha$ was also included as a hyperparameter.

All models that are based on the \drsa{} architecture, namely \drsa{}, \kamran{}, and \dcs{}, share the following tunable hyperparameters. The \emph{encoder} part represents the dense network structure before the LSTM, the \emph{decoder} the LSTM structure itself and \emph{aggregation} the dense network that converts each of the decoder outputs into a scalar value. Each subnetwork's number of layers and number of nodes were hyperparameter tuned. Moreover, we introduce the observation window \tmax{} that corresponds to the maximum survival time in months for each training dataset. We then use the data-driven approach of defining the total number of output nodes $L$ as a hyperparameter based on \tmax{}, such that $L \in \{.25, .5, 1, 2\} \cdot \text{\tmax{}}$. Note that this approach only sets the number of output nodes $L$ of the network and is independent of the temporal spacing (linear, logarithmic or quantile) of those nodes. Furthermore, instead of fixing the loss weighting parameter $\lambda$ to $1$ as in the original \kamran{} model, it is also included together with $\sigma$ \eqref{eq:ours:lkernel} in the hyperparameter search space for \kamran{} and all \dcs{} models.\\

Table \ref{tab:htune:best} shows the best hyperparameters per model for each dataset. For all discrete-time models, the optimal network architectures tend to be rather shallow. In some cases, some parts of the architecture, namely encoder, decoder, or aggregation are dismissed. Notice that \deepsurv{} and \coxtime{} almost always have best performance with shallow neural networks -- usually with one hidden layer, even though the search space included up to five. For the aggregation part of the network, one fully connected layer is necessary to map the decoder's output to a single hazard rate $h_l$. In our approach, we allow additional fully connected aggregation layers for a deeper model architecture. We also analyzed the influence of the total number of output nodes in the hyperparameter tuning.  Here we can see incoherent results that might depend on the dataset as well as the temporal spacing of output nodes. In some cases, increasing the number of output nodes yields the best results (e.g. DCS-linear, DCS-log on \support{} and \metabric{}), in others (e.g. DCS-quant on \support{} and \metabric{}) less total output nodes $L$ were selected.

\begin{table*}[ht] % hyperparams
\begin{center}
\caption{Best hyperparameters for all analyzed datasets and the presented models.}
\resizebox{\linewidth}{!}{%
\begin{tabular}{lrrrrrrr}
\textbf{\support{}} & DeepSurv & CoxTime & \drsa{} & \kamran{} & DCS-linear & DCS-log &  DCS-quant \\
\midrule
encoder\_num\_layers & 1 & 1 & 0 & 0 & 1 & 2 & 2 \\
encoder\_nodes\_per\_layer & 64 & 32 & - & - & 32 & 128 & 64 \\ \midrule
decoder\_num\_layers & & & 1 & 1 & 1 & 0 & 1\\
decoder\_nodes\_per\_layer & & & 128 & 32 & 32 & - & 64 \\
decoder\_bidirectional & & & & & 0 & - & 1  \\
decoder\_use\_lstm\_skip & & & & & 0 & - & 0   \\ \midrule
aggregation\_num\_layers & & & & 1 & 1 & 1 & 2 \\
additional\_aggregation\_nodes\_per\_layer & & & & - & - & - & 32  \\
output\_grid\_num\_nodes & & & 280 & 67 & 140 & 140 & 35  \\\midrule
$\alpha$ & & & 0.36 & & & & \\
$\lambda$ & & & & 2.00 & 0.85 & 0.25 & 2.00 \\
$\sigma$ & & & & 1.08 & 1.55 & 2.00 & 1.00 \\
\\
\textbf{\metabric{}} & DeepSurv & CoxTime & \drsa{} & \kamran{} & DCS-linear & DCS-log &  DCS-quant \\ \midrule
encoder\_num\_layers & 1 & 1 & 0 & 0 & 2 & 1 & 0 \\
encoder\_nodes\_per\_layer & 8 & 8 & - & - & 128 & 64 & -\\ \midrule
decoder\_num\_layers & & & 1 & 1 & 2 & 1 & 2 \\
decoder\_nodes\_per\_layer & & & 128 & 64 & 64 & 64 & 32 \\
decoder\_bidirectional & & & & & 0 & 1 & 0 \\
decoder\_use\_lstm\_skip & & & & & 1 & 1 & 0 \\ \midrule
aggregation\_num\_layers & & & & 1 & 2 & 1 & 1 \\
additional\_aggregation\_nodes\_per\_layer & & & & - & 32 & - & - \\
output\_grid\_num\_nodes & & & 350 & 350 & 700 & 350 & 60 \\ \midrule
$\alpha$ & & & 0.03 & & & & \\
$\lambda$ & & & & 1.94 & 1.19 & 1.08 & 0.25 \\
$\sigma$ & & & & 1.63 & 0.25 & 1.45 & 2.00 \\
\\
\textbf{\flchain{}} & DeepSurv & CoxTime & \drsa{} & \kamran{} & DCS-linear & DCS-log &  DCS-quant \\ \midrule
encoder\_num\_layers & 5 & 1 & 1 & 0 & 1 & 0 & 1 \\
encoder\_nodes\_per\_layer & 64 & 8 & 64 & - & 64 & - & 64 \\ \midrule
decoder\_num\_layers & & & 1 & 1 & 1 & 1 & 0 \\
decoder\_nodes\_per\_layer & & & 16 & 32 & 64 & 128 & - \\
decoder\_bidirectional & & & & & 1 & 1 & - \\
decoder\_use\_lstm\_skip & & & & & 1 & 0 & - \\ \midrule
aggregation\_num\_layers & & & & 1 & 1 & 1 & 1 \\
additional\_aggregation\_nodes\_per\_layer & & & & - & - & - & - \\
output\_grid\_num\_nodes & & & 42 & 171 & 125 & 42 & 85 \\ \midrule
$\alpha$ & & & 0.39 & & & & \\
$\lambda$ & & & & 1.34 & 2.00 & 0.67 & 1.13 \\
$\sigma$ & & & & 0.34 & 1.00 & 0.46 & 0.73 \\
\end{tabular}
}
\label{tab:htune:best}
\end{center}
\end{table*}

\section{Event histograms of other datasets}
\label{apx:datasets:histograms}

\begin{figure}[hb] % figure grids
    \centering
    
    \subfigure[\metabric{}]
    {
        \includegraphics[width=\linewidth]{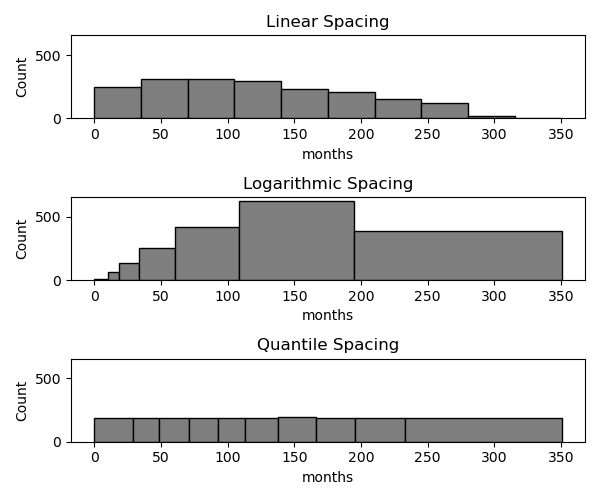}
        \label{fig:output_grid_metabric}
    }
    
    \subfigure[\flchain{}]
    {
        \includegraphics[width=\linewidth]{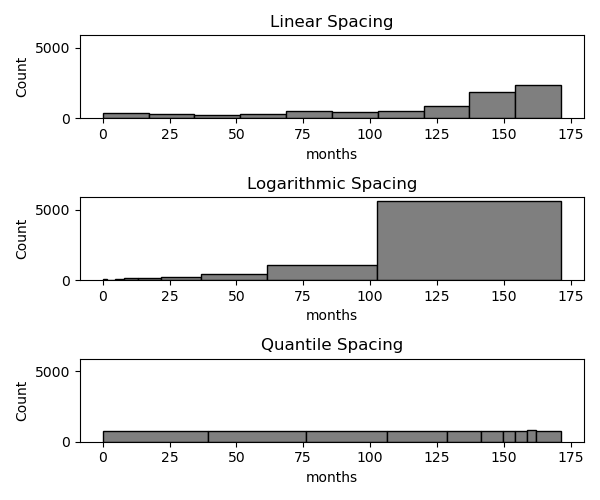}
        \label{fig:output_grid_flchain}
    }    
    
    \label{fig:output_grids:rest}
    \caption{Histograms of the number of events per temporal prediction on \metabric{} and \flchain{}. Ten buckets in time are made with linear, logarithmic and quantile temporal spacing.}
\end{figure}

Histograms of the number of events per temporal prediction for the \metabric{} and \flchain{} datasets are shown in \figureref{fig:output_grids:rest}.
 
% \begin{table*}[t] % no. comparison
% \begin{center}
% \caption{Basic properties of the evaluated survival models regarding linear or non-linear feature encoding, if the Cox proportionality assumption must be accounted for (prop. assump.), if the output is continuous or discrete (output), if the model uses a relative or absolute objective function (obj. func.), if the model emphasized calibration (emph. calib.) and if event vs. censoring comparisons are included (EC comp.).}
% % \resizebox{\linewidth}{!}{%

% \begin{tabular}{rlrrrrll}
%  &  & \textbf{feature encoding} & \textbf{prop. assump.} & \textbf{output} & \textbf{obj. func.} & \textbf{emph. calib.} & \textbf{EC comp.} \\
%  \toprule
% \cite{Cox1972} & CoxPH & linear & yes & continuous & relative & no & yes \\
% \cite{Katzman2018} & DeepSurv & non-linear & yes & continuous & relative & no & yes \\
% \cite{Kvamme2019} & CoxTime & non-linear & no & continuous & relative & no & yes \\
% \cite{Ren2019} & \drsa{} & non-linear & no & discrete & absolute & no & no \\
% \cite{Kamran2021} & \kamran{} & non-linear & no & discrete & absolute + relative & yes & no \\
% {}    [Ours] & \textbf{DCS} & non-linear & no & discrete & absolute + relative & yes & yes
% \end{tabular}

% \label{table:model:comparison}
% \end{center}
% \end{table*}